\documentclass[11pt,a4paper]{article}
\usepackage[hyperref]{emnlp2020}
\usepackage{times}
\usepackage{latexsym}

\usepackage{microtype}

\aclfinalcopy 
\def\aclpaperid{213} 

\usepackage{amsmath}
\usepackage{multirow}
\usepackage{amssymb}
\usepackage{graphicx}

\title{FastTrees: Parallel Latent Tree-Induction for Faster Sequence Encoding}

\author{
Bill Tuck Weng Pung*, Alvin Chan \\
School of Computer Science and Engineering, Nanyang Technological University \\
*\texttt{pung0011@e.ntu.edu.sg} \\ 
}

\date{}

\begin{document}
\maketitle
\begin{abstract}
Inducing latent tree structures from sequential data is an emerging trend in the NLP research landscape today, largely popularized by recent methods such as Gumbel LSTM and Ordered Neurons (ON-LSTM). This paper proposes \textsc{FastTrees}, a new general-purpose neural module for fast sequence encoding. Unlike most previous works that consider recurrence to be necessary for tree induction, our work explores the notion of parallel tree-induction, i.e., imbuing our model with hierarchical inductive biases in a parallelizable, non-autoregressive fashion. To this end, our proposed \textsc{FastTrees} achieves competitive or superior performance to ON-LSTM on four well-established sequence modeling tasks, i.e., language modeling, logical inference, sentiment analysis and natural language inference. Moreover, we show that the \textsc{FastTrees} module can be applied to enhance Transformer models, achieving performance gains on three sequence transduction tasks (machine translation, subject-verb agreement and mathematical language understanding), paving the way for modular tree-induction modules. Overall, we outperform existing state-of-the-art models on logical inference tasks by $+4\%$ and mathematical language understanding by $+8\%$. Code is available on Github\footnote{\url{https://github.com/billptw/FastTrees}}.
\end{abstract}

\section{Introduction}

Inducing and imposing hierarchical tree-like inductive biases in sequential models has garnered increasingly significant attention \cite{shen2020structformer,shen2018ordered,shen2017neural,yogatama2016learning,jacob2018learning,havrylov2019cooperative}, largely owing to the promise of automatically capturing intrinsic syntactic and linguistic structures prevalent in many forms of sequential data (e.g., language, mathematics and music). This is reflected in recent works, i.e., Ordered Neurons (ON-LSTM \cite{shen2018ordered}) and Gumbel LSTM \cite{choi2018learning}, which have shown that imbuing sequential models with hierarchical inductive biases is a fruitful endeavor. After all, learning with well-suited architectural inductive biases generally improves representation learning.

While most prior work relies on autoregressive methods for tree induction \cite{shen2018ordered,choi2018learning}, this paper investigates the notion of a separable, modular tree-induction module which learns to induce latent trees in a parallelizable, non-autoregressive fashion. The key idea is to de-couple tree induction from the recursive loop and parameterize the tree induction module with parallelizable methods, such as convolution or position-wise feed-forward layers, for efficient computation on GPU-enabled machines. To this end, the tree-induction module no longer relies on the hidden-to-hidden transition. The key motivation is that tree composition decisions are not necessarily global (i.e., access to all previous tokens via a compressed hidden state memory may not be necessary). Conversely, learning from local neighborhoods may generate higher performing induced trees.

In light of the inherent speed and efficiency benefits, we coin our proposed methods \textsc{FastTrees}. Empirically, we show that parallel tree induction is highly effective in speed and performance, with $20\%-30\%$ faster inference than the recent ON-LSTM while achieving competitive performance on a suite of NLP tasks. Moreover, we find that parallel tree induction shines in processing formal language, exceeding the state-of-the-art performance by $+4\%$ (absolute percentage) on the logical inference task \cite{bowman2015tree}. On the unsupervised parsing task, we show that \textsc{FastTrees} is capable of producing high fidelity trees, with competitive performance when compared to the autoregressive tree induction method in ON-LSTM. Additionally, we move beyond parallel tree induction and propose \textsc{Faster FastTrees}, a quasi-recurrent model that dispenses its reliance on hidden-to-hidden transitions, achieving $\approx40\%$ speed gain over ON-LSTM. 

Aside from its enhanced inference speed over the ON-LSTM, the feasibility of non-autoregressive tree induction opens up new avenues of research on modular tree-induction modules. Beyond just faster ON-LSTM units, we propose a new general-purpose \textsc{FastTrees} module that learns to induce trees on a sequence-level. This allows us to learn tree-structured gating functions for Transformer models \cite{vaswani2017attention}. Our experiments show the efficacy of the \textsc{FastTrees}-enhanced Transformers, ameliorating vanilla Transformers performance on a variety of tasks like machine translation, subject-verb agreement, and even attaining $+8\%$ absolute improvement in accuracy for mathematical language understanding.

\paragraph{Our contributions} Overall, the prime contributions of this work can be summarized as follows:
\begin{itemize}
    \item We propose that tree induction can be made parallel and non-autoregressive. We show the general effectiveness of parallel trees on a variety of NLP tasks, demonstrating that recurrence is not necessary for tree induction.
    \item We propose \textsc{FastTrees}, adapted from the recent state-of-the-art ON-LSTM \cite{shen2018ordered} for improved efficiency. Our proposed \textsc{FastTrees} outperforms ON-LSTM on sentiment analysis, natural language inference and logical inference tasks. This is achieved while enjoying $20\%-40\%$ in speed gains. Additionally, we propose an even faster variation of \textsc{FastTrees} that completely dispenses with hidden-to-hidden transitions altogether, also achieving similar and competitive performance across several tasks.
    \item We propose \textsc{FastTrees} Transformer models that outperform regular Transformers on Mathematical Language Understanding tasks by $+8\%$.
\end{itemize}

\section{Related Work}
Learning to induce hierarchical structures from sequential data has shown immense potential in many recent works \cite{shen2018ordered,choi2018learning,drozdov2019unsupervised,shen2017neural,bowman2015tree,jacob2018learning}. After all, many forms of sequential data, especially language, are intrinsically hierarchical in nature. Latent tree induction can potentially benefit the representation learning process, especially since finding suitable inductive biases forms the cornerstone of deep learning research.

Recurrent neural networks (RNN) and its variants (e.g., LSTMs \cite{hochreiter1997long} and/or GRUs \cite{cho2014learning} have been highly effective inductive biases for reasoning with sequences. Over the years, there had been many algorithmic innovations and improvements to the recurrent unit. A clear and promising line of research is to incorporate recursive structures and trees into sequence models. An early work, the Tree-LSTM \cite{tai2015improved}, explicitly composes sequences by leveraging syntactic information, just like many of the predecessor work, e.g., recursive neural networks \cite{socher2013recursive}.  Subsequently, \cite{bowman2016fast} proposed a stack-augmented neural network trained by syntax information.

Learning latent tree structures without explicit syntactic supervision has demonstrated recent success. \cite{shen2017neural} proposed Parse-Read-Predict  Network (PRPN), a model that leverages self-attention and learned syntactic distance to induce latent trees. Similarly, the Gumbel LSTM \cite{choi2018learning} learns the discrete composition of tokens in a sequence using the Gumbel Softmax operator. Learning to compose with reinforcement learning is also a prominent approach \cite{yogatama2016learning}. Recently, \cite{havrylov2019cooperative} proposed to jointly learn syntax and semantics via Proximal Policy Optimization (PPO).

The most directly relevant work, the ON-LSTM \cite{shen2018ordered} proposed a new activation function (cumulative Softmax) in order to imbue ordered hierarchies in the gating functions of the LSTM. The tree-inducing gating functions are parameterized in a similar fashion to the main LSTM unit and are also conditioned on the previous hidden state. 

The landscape of sequence encoding has recently shifted towards parallel paradigms, instead of relying on the relatively slower, step-by-step recurrence. This is made notable by the inception of models such as the Transformer \cite{vaswani2017attention} or Convolutional models \cite{wu2019pay}. This work investigates the relative importance of the autoregressive induction of tree structure. Our work is closely related to recent works that attempt to parallelize the recurrent unit, e.g., Quasi-Recurrent Networks \cite{bradbury2016quasi} or Simple Recurrent Units \cite{lei2017training}. To this end, this work brings forward a novel perspective of investigating the nature of autoregressive versus non-autoregressive models within the context of latent tree-induction.

\section{Our Proposed Method}
This section describes our proposed method. Our key innovation lies in proposing sequence encoders that typically accept $\ell$ vectors of $d$ dimensions. For recurrent models, the recursive loop is defined as $h_{t}, c_{t} = \text{RNN}(x_{t}, h_{t-1}, c_{t-1}) $.
The output representation at each time step is the hidden state $h_{t}$. On the other hand, $c_{t}$ is the internal state of the RNN unit at timestep $t$.

\begin{figure*}[ht]
    \centering
    \resizebox{0.65\paperwidth}{!}{
    \includegraphics[]{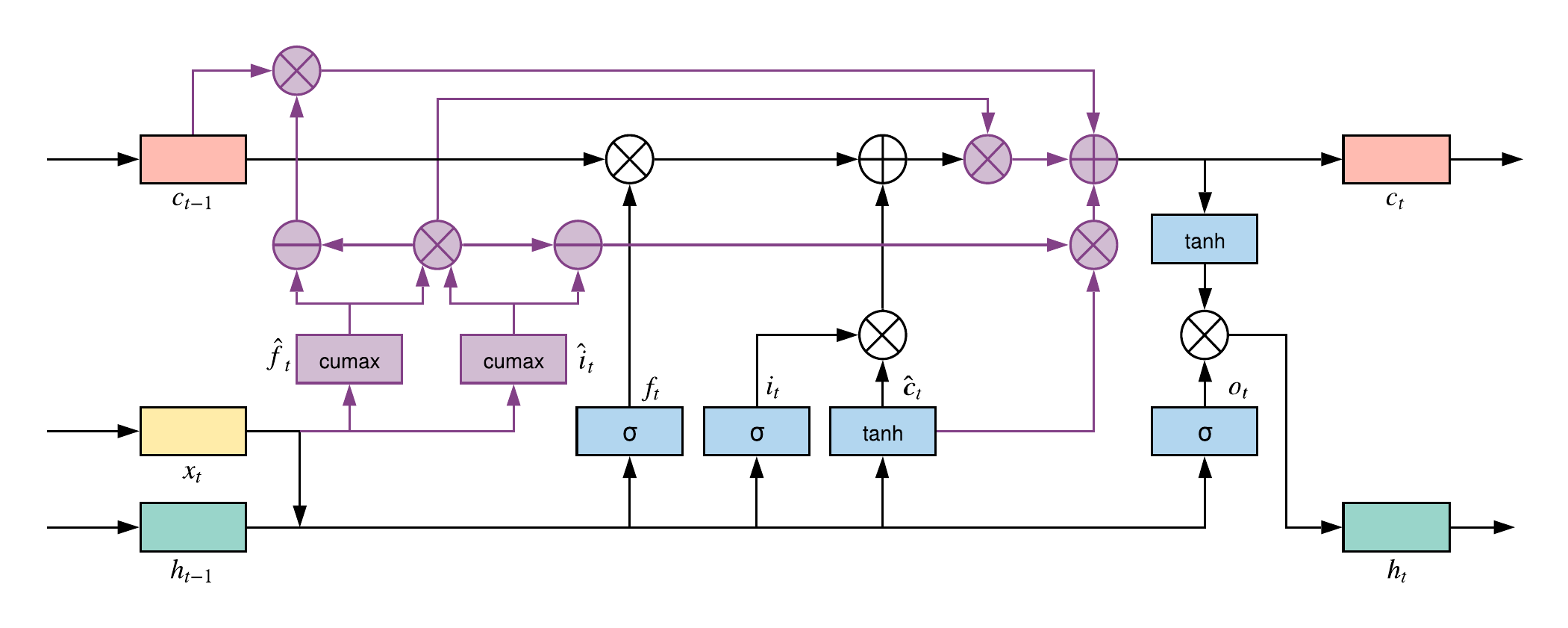}
    }
    \caption{\textsc{FastTrees} model, showing the parallel tree-induction module (in purple) in an unrolled LSTM unit (vanilla LSTM connections in black). The master forget gate $\hat{f}_{t}$ and master input gate $\hat{i}_{t}$ now rely only on the input of the current time step ($x_t$), and thus can be computed in parallel.}
    \label{fig:model}
\end{figure*}

\subsection{\textsc{FastTrees}}
Our \textsc{FastTrees} unit accepts an input sequence of $X \in \mathbb{R}^{\ell \times d}$ where $\ell$ is the length of the sequence and $d$ is the input dimension. \textsc{FastTrees} can be interpreted as a modification of the standard LSTM cell. As a necessary exposition within the context of this work, the details of the standard LSTM cell are described as follows:
\begin{align*}
f_{t}  &= \sigma(W_fx_t + U_fh_{t-1} + b_f) \\
i_{t}  &= \sigma(W_ix_t + U_ih_{t-1} + b_i) \\
o_{t}  &= \sigma(W_ox_t + U_oh_{t-1} + b_o) \\
\hat{c}_{t} &= tanh(W_cx_t + U_ch_{t-1} + b_c) \\
c_{t} &= f_t \odot c_{t-1} + i_{t} \odot \hat{c}_{t} \\
h_{t} &= o_{t} \odot tanh(c_{t})
\end{align*}
where $x_{t}$ is the input token at timestep $t$. $h_{t}$ is the hidden state of the LSTM unit at timestep $t$. $W_{*}, U_{*}, b_{*}$ where $*=\{f,i,o,c\}$ are the parameters of the unit. $\sigma$ is the sigmoid activation function and $\odot$ is the element-wise (Hadamard) product.  Gating functions are parameterized by linear transformations applied on both the input token and previous hidden state. Our modification lies in the construction of structured, hierarchical gating functions, described as follows:

\paragraph{Parallel Tree-Induction} In order to learn tree-structures in a non-autoregressive fashion, we compute the master forget gate $\hat{f}_{t}$ and master input gate $\hat{i}_{t}$ as follows:
\begin{align*}
\hat{f}_{t} &= \text{cumax}(F_f(x_{t}))_{t} \\ 
\hat{i}_{t} &= 1 - \text{cumax}(F_i(x_{t}))_{t}
\end{align*}
where $F_f(.)$ and $F_i{}(.)$ are parameterized functions. cumax() is the cumulative Softmax function which essentially applies the cumulative sum function right after a Softmax operator. Note that this is in contrast with ON-LSTM \cite{shen2018ordered} which considers $\hat{f}_{t} = \text{cumax}(W_{\hat{f}}x_{t} + U_{\hat{f}}h_{t-1} + b_{\hat{f}})$ and  $\hat{i}_{t} = 1 -\text{cumax}(W_{\hat{i}}x_{t} + U_{\hat{i}}h_{t-1} + b_{\hat{i}})$. Instead, our formulation allows parallel computation of $\hat{f}_t$ and $\hat{i}_t$. The standard choice of $F_f(.)$ and $F_i{}(.)$ are 2 layered position-wise feed-forward layers:
\begin{align*}
F(x_{t}) = W{x_{t}} + b_{t}
\end{align*}
Alternatively, we may also consider incorporating local information into the tree induction mechanism using causal 1D convolutions followed by a single positional feed-forward layer.
\begin{align*}
F(X) = F_{P}(\text{CausalConv}(X))
\end{align*}
In this case, the sequence token $x_{t}$ should not have access to tokens $>t$. $F_{P}(.)$ is a position feed-forward layer. The two variants are named \textsc{FastTrees} and Conv. \textsc{FastTrees} respectively. 

Next, to learn tree-structured hidden representations, the structured gating mechanism is defined as follows:
\begin{align*}
\omega_t &= \hat{f}_{t} \odot \hat{i}_{t} \\    
\hat{f}_{t} &= f_{t} \odot \omega_t + (\hat{f}_{t} - \omega_{t}) \\
\hat{i}_{t} &= i_{t} \odot \omega_t + (\hat{i}_{t} - \omega_{t}) \\
c_{t} &= \hat{f}_t \odot c_{t-1} + \hat{i}_t \odot \hat{c}_{t}
\end{align*}
where $c_{t}$ is the new cell state at time step $t$.
The gates are now imbued with a hierarchical structure, due to the splitting of points defined by the cumax function. We refer interested readers to \cite{shen2018ordered} for more details.

\subsection{\textsc{Faster FastTrees}}
We move beyond parallel tree induction and propose an even faster variant of \textsc{FastTrees}. We completely remove the reliance of the hidden-to-hidden transition within the recurrent unit. The \textsc{Faster FastTrees} unit is described as follows:
\begin{align*}
f_{t}&=\sigma(F_f(x_{t})),\: i_{t}=\sigma(F_i(x_{t})), \: o_{t}=\sigma(F_o(x_{t})) \\
\hat{f}_{t} &= \text{cumax}(F_f(x_{t}))_{t},\:\: \hat{i}_{t} = 1 - \text{cumax}(F_i(x_{t}))_{t} \\
\omega_t &= \hat{f}_{t} \odot \hat{i}_{t} \\  
\hat{f}_{t} &= f_{t} \odot \omega_t + (\hat{f}_{t} - \omega_{t}), \:\: \hat{i}_{t} = i_{t} \odot \omega_t + (\hat{i}_{t} - \omega_{t}) \\
\hat{c}_{t} &= tanh(W_cx_t + b_c), \:\: c_{t} = \hat{f}_t \odot c_{t-1} + \hat{i}_{t} \odot \hat{c}_{t} \\
h_{t} &= o_{t} \odot tanh(c_{t})
\end{align*}
where $F_f(x_{t}),F_i(x_{t}),F_o(x_{t})$ are linear transformations. This formulation is similar in spirit to QRNNs \cite{bradbury2016quasi} and can be interpreted as the quasi-recurrent adaptation of ON-LSTM. Naturally, this formulation gains speed, as more gate construction functions are now parallelizable.

\begin{table*}
    \centering
    \begin{tabular}{lcccc}
    \hline
    Model & Parameters & Val ppl & Test ppl & Time (sec) \\
    \hline
    PRPN-LM \cite{shen2017neural} & - & - & 62.00 & - \\
    4-Layer Skip Connection \cite{melis2017state} & 24M & 60.90 & 58.30 & - \\
    AWD-LSTM \cite{merity2017regularizing} & 24M & 60.00 & 57.30 & - \\
    ON-LSTM \cite{shen2018ordered} & 25M & \textbf{58.32} & \textbf{56.25} & 296 \\
    \hline
    Faster \textsc{FastTrees} (This work) & 25M & 61.93 & 59.10 & \textbf{173} $(+42\%)$ \\
    Conv. \textsc{FastTrees} (This work) & 25M & 59.88 & 57.30 & 204 $(+31\%)$  \\
    \textsc{FastTrees} (This work) & 25M & 58.47 & 56.35 &  209 $(+29\%)$ \\
    \hline
    \end{tabular}
    \caption{Model perplexity (lower is better) on validation and test sets for language modeling task on PTB. Train times measured by wall-clock in seconds per epoch. Percentage speed improvements of \textsc{FastTrees} over ON-LSTM denoted in parenthesis.} 
    \label{tab:lm}
\end{table*}

\subsection{\textsc{FastTree} Transformers}
To demonstrate the general-purpose utility of our method, we show how our \textsc{FastTrees} module can be added to boost the performance of Transformers models. The key idea is to show that this hierarchical inductive bias can be useful when added to other sequence transduction models. This module accepts an input $X \in \mathbb{R}^{\ell \times d}$ and produces an output representation of equal shape $Y \in \mathbb{R}^{\ell \times d}$. Note that we switch to matrix notation since our operations now operate at sequence-level.
\begin{align*}
A &= \text{cumax}(F_{a}(X)) \\
B &= (1 - \text{cumax}(F_{b}(X))) \\
F &= ((\sigma(F_{h}(X))) \odot (A \odot B)) + (A - (A \odot B)) \\
Y & = F \odot X
\end{align*}
$F_{a}(.), F_{b}(.), F_{h}(.)$ are parameterized functions which may be sequence-level operations (e.g., convolution or position-wise feed-forward layers). The key idea behind this approach is to convert the token-level ordered hierarchical inductive bias \cite{shen2018ordered} into sequence level operations. The interpretation of \textsc{FastTrees} Transformer is reminiscent of the $\hat{f},\hat{i}$ master gating functions in ON-LSTM and/or \textsc{FastTrees}. Here, we simply learn a single structured forget gate, directly masking the output representations of this layer in a tree-structured fashion.
\paragraph{Overall Architecture}
We place the tree induction module denoted \textsc{FastTrees}(.) right after the self-attention layer. A single Transformer block is now written as:
\begin{align*}
\hat{X} &= \frac{1}{\sqrt{d}}\:\text{Softmax}(W_{Q}(X')W_{K}(X'))^{\top}(W_{V}(X') \\
Y' &= \textsc{FastTrees}(\hat{X}) \\ 
\hat{Y} &= F_{\phi}(Y'))
\end{align*}
where $F_{\phi}(.)$ is a two-layered position-wise feed-forward network with ReLU activations. Notably, our extension inherits all other properties from the base Transformer, including its multi-headed and multi-layered nature.

\section{Experiments}
We evaluate \textsc{FastTrees} units on four well-established sequence modeling tasks (language modeling, logical inference, sentiment analysis and natural language inference), and \textsc{FastTrees}-augmented Transformers on three tasks (machine translation, subject-verb agreement and mathematical language understanding). All timed comparisons
were tested by training the models on NVIDIA Tesla V100 GPUs for fair comparison, and averaging the wall clock time per epoch over several cycles.

\subsection{Word-level Language Modeling}
Word-level language modeling is an important test of the modeling capabilities of a neural network in various linguistic phenomena. In evaluating our model's representational abilities, we test it by measuring perplexity on the Penn Treebank (PTB) \cite{marcus-etal-1993-building} dataset \footnote{\url{https://catalog.ldc.upenn.edu/LDC99T42}}.

\paragraph{Experimental Setup} To evaluate our model performance fairly, we follow the model hyper-parameters, regularization and optimization techniques used in ON-LSTM \cite{shen2018ordered}. We use a three-layer model with 1150 units in the hidden layer, and an embedding size of 400. We apply dropout on the word vectors, the output between LSTM layers, the output of the final LSTM layer, and embedding dropout of (0.5, 0.3, 0.45, 0.1) respectively. We perform grid search on hidden layer sizes in the range of [600,1500] with the step size of 50, and found 1150 to have the best validation performance.


\paragraph{Results} Table \ref{tab:lm} reports our results on the word-level language modeling task. Keeping the number of layers, hidden state and embedding dimensions constant, we show that the performance attained by \textsc{FastTrees} is competitive to ON-LSTM while training $29-42\%$ faster, and also outperforms AWD-LSTM \cite{merity2017regularizing}.

\begin{table*}
    \centering
    \begin{tabular}{lccccccc}
    \hline
     & \multicolumn{6}{c}{\% Accuracy on each sequence length} &  \\
    Model                          & 7  & 8  & 9  & 10 & 11 & 12 & Time (sec) \\
    \hline
    TreeLSTM$^\dagger$ \cite{tai2015improved} & 94.0 & 92.0 & 92.0 & 88.0 & 97.0 & 86.0 & - \\
    \hline
    LSTM                           & 88.0 & 84.0 & 80.0 & 78.0 & 71.0 & 69.0 & - \\
    RRNet \cite{jacob2018learning} & 84.0 & 81.0 & 78.0 & 74.0 & 72.0 & 71.0 & - \\
    ON-LSTM \cite{shen2018ordered} & 91.0 & 87.0 & 85.0 & 81.0 & 78.0 & 75.0 & 85 \\
    \hline
    Faster \textsc{FastTrees} & 66.0 & 62.0 & 57.0 & 55.0 & 53.0 & 53.0 & \textbf{49.0} $(+42\%)$ \\
    \textsc{FastTrees} & 91.0 & 88.0 & 83.0 & 80.0 & 76.0 & 74.0 & 75.0 $(+11\%)$ \\
    Conv. \textsc{FastTrees} & \textbf{93.0} & \textbf{90.0} & \textbf{86.0} & \textbf{83.0} & \textbf{80.0} & \textbf{79.0} & 58.0 $(+32\%)$  \\
    \hline
    Accuracy Gain (abs.) over ON-LSTM  & $+2\%$ & $+3\%$ &$+1\%$ &$+2\%$ & $+2\%$ & $+4\%$\\
    \hline
    \end{tabular}
    \caption{Percentage accuracy on test set of models on each sequence length (from 7 to 12) for logical inference task. $\dagger$ denotes with ground truth syntax. Train times measured by wall-clock in seconds per epoch. Percentage speed improvements of \textsc{FastTrees} over ON-LSTM denoted in parenthesis.} 
    \label{tab:logic}
\end{table*}

\subsection{Logical Inference}
The logical inference task \cite{bowman2015tree} tests for a model's ability to exploit a recursively defined language to generalize sentences with complex unseen structures. The key idea is that models with appropriate hierarchical inductive bias will do well on this task. 

Sentences in this language use a combination of up to six word types $(p_1, p_2, p_3, p_4, p_5, p_6)$ and three logical operations $(and, or, not)$. The length of a sentence pair (termed as the sequence length) is defined by the number of logical operators within the longer of the two sentences. The relationship between the two sentences are described using seven mutually exclusive logical relations: two directions of entailment $(\sqsubset,\sqsupset)$, equivalence $(\equiv)$, exhaustive and non-exhaustive contradiction $(\wedge,\mid)$, and two types of semantic independence $(\#,\smile )$. The task is to predict the correct logical relationship given a sequence. 

\paragraph{Experimental Setup} We train the model with sequence lengths varying up to 6, while evaluating the model separately on each sequence length (up to 12). Notably, performance on sequence lengths between 7 and 12 is representative of its ability to generalize to unseen sentence structures, given their absence in the training data. We use a 80/20\% train/test split, with 10\% of the training set reserved for the validation set. We parameterize the RNN models with a hidden layer of size 400, and the input embedding of size 128. A dropout of 0.2 was applied between different layers.


\paragraph{Results} Table \ref{tab:logic} illustrates the performance improvements of Conv. \textsc{FastTrees} over ON-LSTM. Conv. \textsc{FastTrees} outperforms ON-LSTM, especially in longer sentence sequences with a $+4\%$ absolute improvement for sequence length 12. This shows its ability to learn recursive data structures while operating $+32\%$ faster. The key idea here is to show that learning trees in a parallel, non-autoregressive fashion can be sufficiently powerful as a model hierarchical structure. 


\begin{table}[ht]
    \centering
    \resizebox{0.95\columnwidth}{!}{
    \begin{tabular}{lccc}
    \hline
    Model & Parameters & Acc (\%) & Time (sec) \\
    \hline
    LSTM & 2.3M & 87.86 & 132 \\
    ON-LSTM & 2.3M & 88.30 & 275  \\
    \hline
    Faster \textsc{FastTrees} & 2.3M & 86.99 & \textbf{161} $(+42\%)$ \\
    Conv. \textsc{FastTrees} & 2.3M & 88.46 & 218 $(+21\%)$ \\
    \textsc{FastTrees} & 2.3M & \textbf{88.69} & 194 $(+30\%)$ \\
    \hline
    \end{tabular}
    }
    \caption{Sentiment classification task with percentage accuracy evaluated on test set of SST-2. Time measured by wall-clock in seconds per training epoch. Percentage speed improvements of \textsc{FastTrees} over ON-LSTM denoted in parenthesis.}
    \label{tab:sst}
\end{table}

\subsection{Sentiment Analysis}
In our experiments, we train our models using the binary Stanford Sentiment Treebank (SST-2) \cite{socher2013recursive}, a 70k sentence dataset where the model predicts a positive or negative sentiment label given an input sentence\footnote{\url{https://www.kaggle.com/atulanandjha/stanford-sentiment-treebank-v2-sst2}}.

\paragraph{Experimental Setup} We initialize the word vectors using two well-established pre-trained embedding types, GloVe 300D \cite{pennington2014glove} and FastText \cite{grave2018learning}. Embedding projections of size $256$ are fed into an encoder with hidden size of $512$, and subsequently an MLP with a hidden layer $D_c $ of size $512$ for the classifier layer. We vary the architecture of the middle encoder layer between ON-LSTM and varying \textsc{FastTrees} models for the experiment. The initial learning rate is set to $0.0004$ and dropped by a factor of $0.2$ when accuracy plateaus, with dropout set to $0.5$. The loss used is standard cross-entropy, and Adam is used for optimization \cite{kingma2014adam}.


\paragraph{Results} High performance in the sentiment analysis task is indicative of hierarchical representation induced, as the successful classification of sentence sentiment is more likely when the model learns a strategy in identifying context and word relations within the entire sentence. From table \ref{tab:sst}, we see how both \textsc{FastTrees} and Conv. \textsc{FastTrees} outperforms ON-LSTM and LSTM on the task, with significant speed-up in training time.  This increase in performance supports our hypothesis of the induced tree representations in \textsc{FastTrees}.

\subsection{Natural Language Inference}
Natural language inference is the task of predicting whether two sentences, a premise sentence and a hypothesis sentence, are neutral, entailing or contradictory. Inference of the latter two is essential in natural language understanding, indicative of a model's ability for semantic representation. We use the Stanford Natural Language Inference (SNLI) dataset \cite{bowman2015large}, which consists of about 570k human-generated, manually-labeled English sentence pairs\footnote{\url{https://nlp.stanford.edu/projects/snli/}}.

\paragraph{Experimental Setup} 
The input to the model is a sentence pair, they are represented individually as the premise sentence vector $h^{pre}$ and the hypothesis sentence vector $h^{hyp}$ by the sentence encoder. The vectors are then concatenated, with embedding projections of size $256$ that are fed into an encoder with hidden size of $512$, and subsequently an MLP with a hidden layer of size $1024$ for the classifier layer. We vary the architecture of the middle encoder layer between ON-LSTM and each \textsc{FastTrees} model for the experiment. The initial learning rate is set to $0.0004$ and dropped by a factor of $0.2$ when accuracy plateaus, with dropout set to $0.2$. The MLP classifier has a dropout of $0.2$. The loss used is standard cross-entropy, and Adam is used for optimization.


\begin{table}[ht]
    \centering
    \resizebox{0.95\columnwidth}{!}{
    \begin{tabular}{lccc}
    \hline
    Model & Parameters & Acc (\%) & Time (sec) \\
    \hline
    LSTM & 4.9M &  86.12  & 657 \\
    Gumbel Tree-LSTM & 10.3M & 86.00 & - \\
    ON-LSTM & 4.9M & 85.82 & 1348 \\
    \hline
    Faster \textsc{FastTrees} & 4.9M & 85.21 & \textbf{850} $(+37\%)$ \\
    \textsc{FastTrees} & 4.9M & 86.04 & 1062 $(+21\%)$ \\
    Conv. \textsc{FastTrees} & 4.9M & \textbf{86.23} & 1052 $(+22\%)$  \\
    \hline
    \end{tabular}
    }
    \caption{Percentage accuracy of models evaluated on test set of SNLI on the natural language inference task. Corresponding size of models in number of parameters denoted in parenthesis. Train times in seconds per epoch generated from wall-clock time. Percentage speed improvements of \textsc{FastTrees} over ON-LSTM denoted in parenthesis.}
    \label{tab:snli}
\end{table}

\paragraph{Results} From table \ref{tab:snli}, we note the improved performance of \textsc{FastTrees} and Conv. \textsc{FastTrees} over ON-LSTM, while training \textgreater$20\%$ faster. Conv. \textsc{FastTrees} also outperforms LSTM and Gumbel Tree-LSTM in this task, with a model size half that of the Gumbel Tree-LSTM.

\subsection{Neural Machine Translation}
 This task involves translating between one language and another. More concretely, we utilize the IWSLT'15 English-Vietnamese (En-Vi) dataset\footnote{\url{https://nlp.stanford.edu/projects/nmt/}}. 

 \paragraph{Experimental Setup} We implement our model with the Tensor2Tensor\footnote{\url{https://github.com/tensorflow/tensor2tensor}} framework, using Transformer Base as the key baseline model. We train both the base Transformer and our proposed models for $50K$ steps using the default hyperparameters. During inference, the length penalty is set to $0.6$ and the beam size is set to $4$. We average the parameters of the last 8 checkpoints.

\paragraph{Results} Table \ref{tab:nmt} illustrates how our proposed method achieves state-of-the-art performance on the IWSLT'15 En-Vi dataset, outperforming not only the base transformer model but also prior work. More importantly, we demonstrate the utility of the \textsc{FastTrees} module, achieving a $1.88\%$ absolute improvement in BLEU score. Both variants of \textsc{FastTrees} are able to enhance the base Transformers model performance in this task.

\begin{table}[ht]
  \centering
  \footnotesize
    \begin{tabular}{lc}
        \hline
         Model & \multicolumn{1}{c}{BLEU}  \\
          \hline
         Luong \& Manning (2015) & 23.30  \\
         Seq2Seq Attention & 26.10 \\
         Neural Phrase-based MT & 27.69\\
         Neural Phrase-based MT + LM & 28.07\\
    Transformer \cite{vaswani2017attention} & 28.43   \\
    \hline
    Transformer +  \textsc{FastTrees} & \textbf{30.31}  \\
    Transformers + Conv. \textsc{FastTrees} & 29.72 \\
    \hline
    Abs. Improvement & +$1.88\%$ \\
    \hline
    \end{tabular}
      \caption{BLEU \cite{papineni2002bleu} scores on machine translation task using IWSLT'15 English-Vietnamese dataset.}
  \label{tab:nmt}
\end{table}

\subsection{Subject-Verb Agreement and Mathematical Language Understanding}
We include additional experiments on subject-verb agreement (SVA) \cite{linzen2016assessing} and mathematical language understanding (MLU) \cite{DBLP:journals/corr/abs-1812-02825} to demonstrate the effectiveness of \textsc{FastTrees}. The SVA task is a binary classification problem, determining if a sentence, \emph{e.g.}, \textit{`The keys to the cabinet \_\_\_\_\_ .'} is followed by a plural or singular verb. This involves language understanding pertaining to learning syntax-sensitive dependencies.  On the other hand, the MLU task involves input sequences such as $x=85, y=-523, x * y$ in which the expected decoding output should be $-44455$.

\paragraph{Experimental Setup} Experiments are conducted on the Tensor2Tensor framework, using the tiny default hyperparameter setting. Models are trained for $10K$ steps for SVA and $100K$ steps for MLU. The evaluation metric is accuracy for the SVA task and accuracy per sequence for MLU.

\begin{table}[ht]
    \footnotesize
    \centering
    \resizebox{0.9\columnwidth}{!}{
    \begin{tabular}{lcc}
    \hline
    Model & SVA & MLU\\
    \hline
    Transformer & 94.8 & 76.10\\
    Transformer + \textsc{FastTrees} & 94.8 & \textbf{84.26}\\
    Transformer + Conv. \textsc{FastTrees} & \textbf{95.7} & 82.62 \\
    \hline
    Abs. Improvement& +$0.9\%$ & +$8.16\%$ \\
    \hline 
    \end{tabular}
    }
    \caption{Accuracy scores on subject-verb agreement prediction task and accuracy per sequence on mathematical language understanding task.}
    \label{tab:add}
\end{table}

\paragraph{Results} Table \ref{tab:add} presents our experimental results on SVA and MLU. Once again, we observe improvements over the base Transformer model. While we observe modest improvements on the SVA task with Conv. \textsc{FastTrees}, the performance gain on the MLU task is very promising. Notably, we achieve an $+8.16\%$ absolute improvement on accuracy per sequence. To this end, we posit that the naturally hierarchical nature of mathematical expressions enables our inductive bias to shine on this task.

\section{Qualitative Analysis}
First, we perform an analysis on the quality of trees produced by models trained on the language modeling tasks. This is determined by the parsing F1 score on the unsupervised constituency parsing task. Second, we visualize the output trees composed by the language models and that of human annotators; these visualizations serve to examine the fidelity of trees produced by \textsc{FastTrees}.

\subsection{Unsupervised Constituency Parsing}
The unsupervised constituency parsing task compares the latent tree structure induced by a pre-trained model with parse outputs annotated by human experts. To ensure reproducibility, we follow the experiment setup proposed in \cite{htut2018grammar}. We first train the models on the language modeling task using the PTB dataset and save those with sufficiently low perplexities. Subsequently, we test them on the Wall Street Journal 10 (WSJ10) dataset and WSJ test set for this task. 

\begin{table}[ht]
    \centering
    \resizebox{0.95\columnwidth}{!}{
    \begin{tabular}{lcccccc} 
        \hline
        \multirow{2}{*}{\textbf{Model}} & \multicolumn{2}{c}{\textbf{Parsing F1}} & \multicolumn{3}{c}{\textbf{ \% Accuracy}} \\
         & \textbf{WSJ10} & \textbf{WSJ} & \textbf{ADJP} & \textbf{NP} & \textbf{PP} \\
        \hline
        Random Trees & 32.2 & 18.6 & 17.4 & 22.3 & 16.0 \\
        Balanced Trees & 43.4 & 24.5 & 22.1 & 20.2 & 9.3 \\
        Left Branching & 19.6 & 9.0 & 17.4 & - & - \\
        Right Branching & 56.6 & 39.8 & - & - & - \\
        \hline
        3300D ST-Gumbel$^\dagger$ & - & 20.1 & 15.6 & 18.8 & 9.9 \\
        w/o Leaf GRU$^\dagger$ & - & 25 & 18.9 & 24.1 & 14.2 \\ 
        300D RL-SPINN$^\dagger$ & - & 13.2 & 1.7 & 10.8 & 4.6 \\
        w/o Leaf GRU$^\dagger$ & - & 13.2 & 1.6 & 10.9 & 4.6 \\
        \hline
        PRPN-LM & \textbf{71.3} & 38.1 & 26.2 & \textbf{63.9} & 24.4 \\ 
        ON-LSTM & 66.8 & \textbf{49.4} & \textbf{46.2} & 61.4 & \textbf{55.4}\\
        \hline
        \textsc{FastTrees} & 65.5 & 44.3 & 43.1 & 54.8 & 52.9 \\
        \hline
    \end{tabular}
    }
    \caption{Unlabeled parsing F1 scores evaluated on full WSJ10 and WSJ test sets. $^\dagger$ denotes models trained on NLI task, and evaluated on full WSJ. The \% Accuracy columns represent percentage of ground truth constituents of a given type that corresponds to constituents in the model parses.}
    \label{tab:parse}
\end{table}

Table \ref{tab:parse} reports the parsing F1 scores obtained from the $2^{nd}$ layer of the 3-layered ON-LSTM and \textsc{FastTrees}. On WSJ10, \textsc{FastTrees} outperforms the random and standard branching baselines, and closely matches ON-LSTM. On WSJ, \textsc{FastTrees} achieves 44.3, outperforming ST-Gumbel \cite{choi2018learning} and RL-SPINN \cite{yogatama2016learning} models and variants. We also note that \textsc{FastTrees} performs well on phrase detection, including adjective phrases (ADJP), noun phrases (NP), and prepositional phrases (PP). From this analysis, we show that with the removal of hidden-to-hidden transition on \textsc{FastTrees} to obtain speed-ups of $20-40\%$, the induced latent trees are still of relatively high fidelity; \textsc{FastTrees} outperforms the standard branching trees, ST-Gumbel, and RL-SPINN models in parsing F1 score, closely resembling the ground truth.

\subsection{Tree Visualization}
\begin{figure}[ht]
    \centering
    \resizebox{0.9\columnwidth}{!}{
    \includegraphics[]{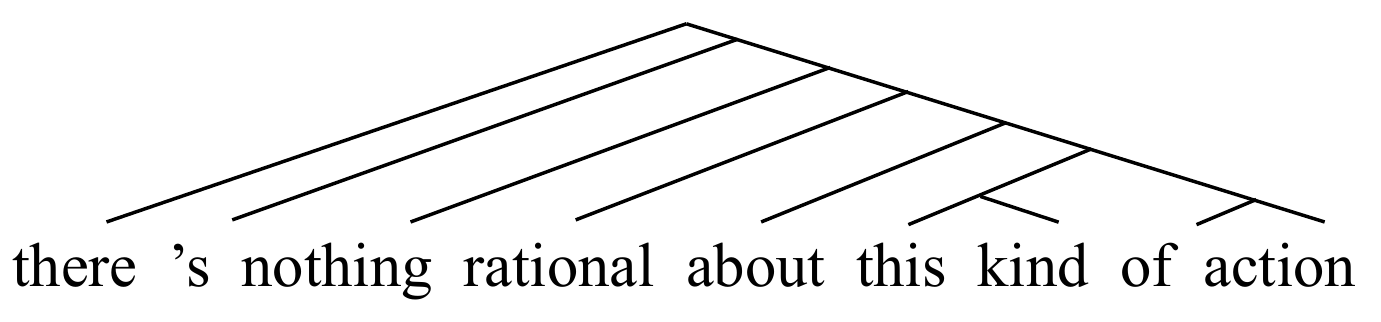}
    \includegraphics[]{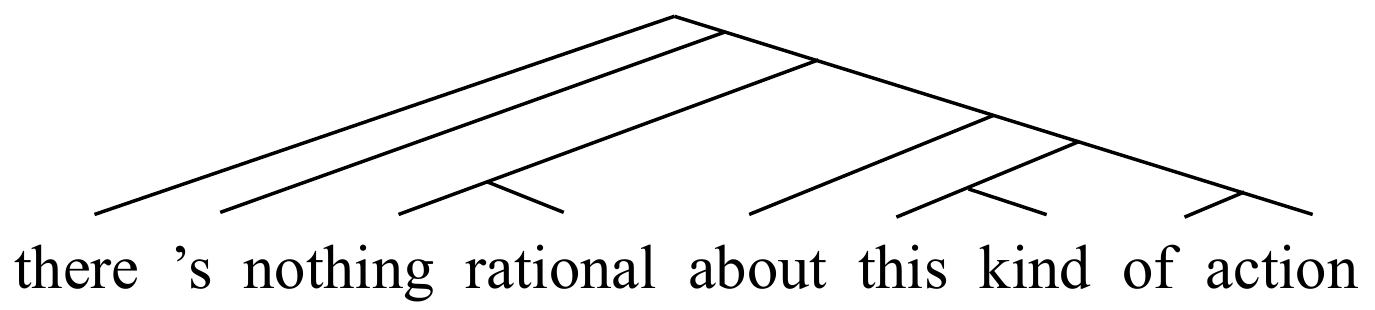}
    }
    \caption{Ground Truth (top), \textsc{FastTrees} output (bottom)}
    \label{fig:rational}
\end{figure}

\begin{figure}[ht]
    \centering
    \resizebox{0.95\columnwidth}{!}{
    \includegraphics[]{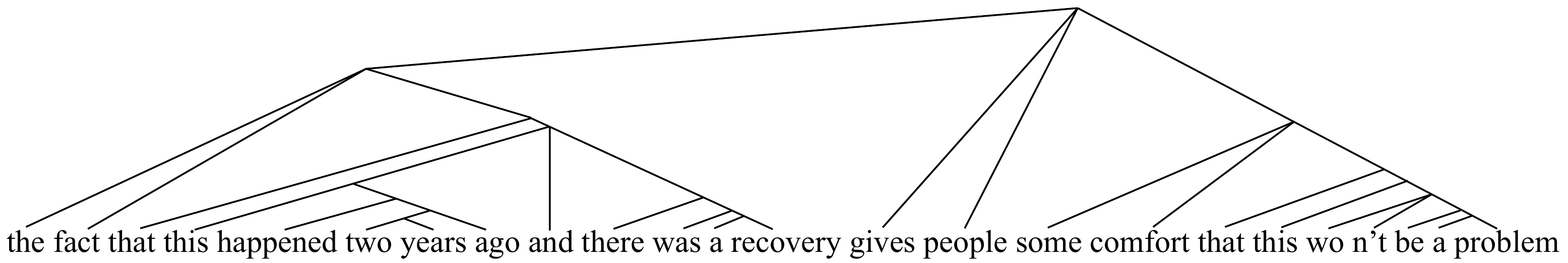}
    }
    \resizebox{0.95\columnwidth}{!}{
    \includegraphics[]{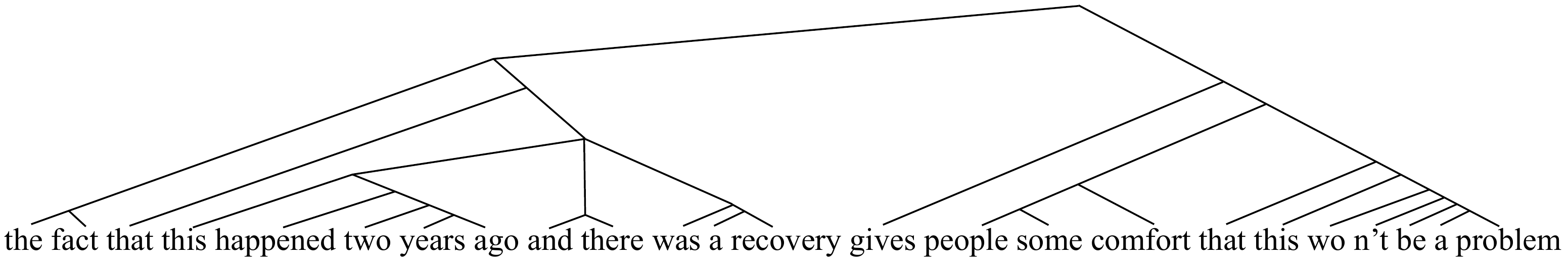}
    }
    \resizebox{0.95\columnwidth}{!}{
    \includegraphics[]{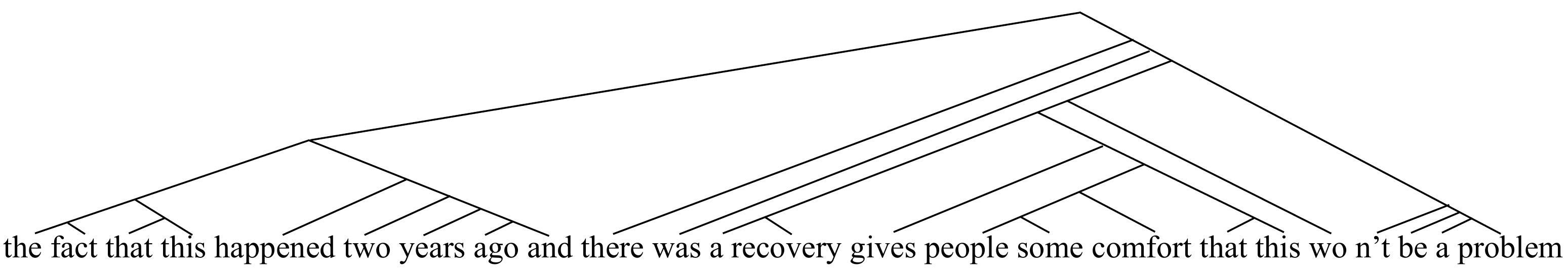}
    }
    \caption{Parse tree from Ground Truth (top), \textsc{FastTrees} (middle), and ON-LSTM \cite{shen2018ordered} (bottom)}
    \label{fig:fact}
\end{figure}

We illustrate samples of the parse output from the unsupervised constituency parsing task in figures  \ref{fig:rational} and \ref{fig:fact}. \textsc{FastTrees} is able to generate parse trees very close to the human-annotated ground truth, as shown in figure \ref{fig:rational}, achieving an F1 score of $85.7\%$. 

Figure \ref{fig:fact} shows how the model output can resemble the ground truth even for longer sequences of $24$ tokens, with an F1 score of $71.0\%$ versus ON-LSTM at $15.4\%$. We surmise that these examples demonstrate the abilities of \textsc{FastTrees} in learning non-trivial tree composition schemes that are useful for many downstream NLP tasks.

\section{Conclusion}
We propose \textsc{FastTrees}, a model that induces latent trees in a parallel, non-autoregressive fashion. Our proposed model outperforms ON-LSTM on logical inference, natural language inference and sentiment analysis while training up to 40\% faster. \textsc{FastTrees} also achieves state-of-the-art results on logical inference, demonstrating that it can learn effectively on intrinsically hierarchical data. Finally, we show that parallel \textsc{FastTrees} can be used to enhance Transformer models, bringing about $8\%$ advancement over MLU.

\bibliographystyle{acl_natbib}
\bibliography{anthology,emnlp2020}

\end{document}